\documentclass{spie}

\usepackage{cite}

\usepackage{times} 
\usepackage{indentfirst} 

\usepackage{amsmath}
\usepackage{amsfonts}
\usepackage{mathtools}
\usepackage{xcolor}
\usepackage{doi}

\usepackage{xkeyval}

\title{Methods of Weighted Combination for Text Field Recognition in a Video~Stream}

\author{Olga Petrova\supit{1,2}, Konstantin Bulatov\supit{1, 2, 3}, Vladimir L. Arlazarov\supit{1, 3}
  \skiplinehalf
  \normalsize 
  \supit{1} Federal Research Center ``Computer Science and Control'' of Russian Academy of Sciences, Moscow, Russia; \\
  \supit{2} Smart Engines Service LLC, Moscow, Russia; \\
  \supit{3} Moscow Institute of Physics and Technology ``MIPT'', Moscow, Russia
}

\begin{document}

\maketitle

\begin{abstract}

  Due to a noticeable expansion of document recognition applicability, there is a high demand for recognition on mobile devices. A mobile camera, unlike a scanner, cannot always ensure the absence of various image distortions, therefore the task of improving the recognition precision is relevant. The advantage of mobile devices over scanners is the ability to use video stream input, which allows to get multiple images of a recognized document. Despite this, not enough attention is currently paid to the issue of combining recognition results obtained from different frames when using video stream input. In this paper we propose a weighted text string recognition results combination method and weighting criteria, and provide experimental data for verifying their validity and effectiveness. Based on the obtained results, it is concluded that the use of such weighted combination is appropriate for improving the quality of the video stream recognition result.
  
\end{abstract}

\section{Introduction}

Optical character recognition systems become widely used in a variety of fields as a part of business, social, and government process automation \cite{a_survey_of_ocr_applications, a_survey_on_ocr_arxiv, 8342338, icmv-jabnoun}. Such systems can take scanned images, photos, and video sequences as an input. The usage of video sequences for documents recognition allows the system to capture the recognized objects under various angles, with various focus, etc. This, in turn, allows to exclude sporadic recognition system errors caused by distortions such as highlights, motion blur, optical system aberrations, etc. \cite{arl-small-scale-cameras, 7881422}. The next step after obtaining per-frame results is to select a strategy of combining these results in a single integrated answer. 

\begin{figure}[ht!]
 \centering
 \includegraphics[width=0.25\textwidth]{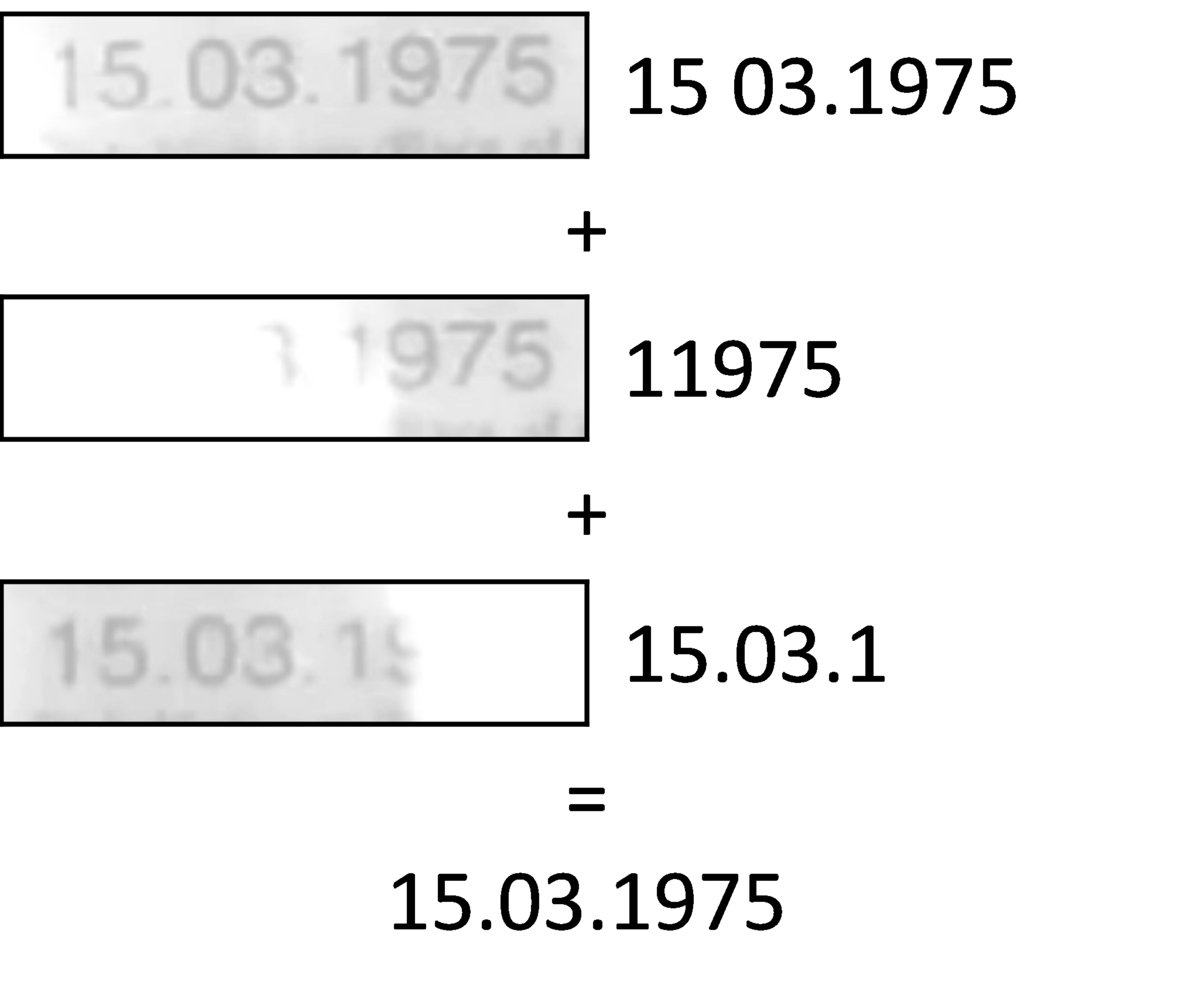}
 \caption{Use case illustration}
 \label{fig:use_case_illustration}
\end{figure}

There are two main groups of approaches to using multiple images of the recognized object for maximization of data extraction efficiency. The first group contains methods and algorithms devoted to combine the information from different frames on an image level, and then to perform recognition of a single image. Such methods include selection of the most informative video frame \cite{6507950}, tracking and combining images of the recognized object \cite{6969211, efficient_video_scene_text_spotting_arxiv}, or using super-resolution methods which allow to merge multiple frames into a single image with higher quality \cite{YUE2016389, 5995614}. The second approach to extract data from video is to combine the recognition results produced independently from each frame \cite{vestnik_integration, chernov2019application}. Figure \ref{fig:use_case_illustration} illustrates an example of combining recognition results from different frames to get the correct answer.

Some parallels can be seen between the task of per-frame recognition results combination and the task of classifier combination (i.e. building an optimal ensemble of multiple recognition systems). For classical classifier combination problem (where the result of each classifier is a class label, possibly with a membership estimation) there exist a large number of combination strategies, such as rules of sum, product, median, maximum, etc. \cite{ensemble_methods, 1688199}. The majority of existing works on such ensembles operate with a single-object classifier model. At the same time, text string recognition is often performed with the string object considered as non-atomic, i.e. consisting of multiple components (characters), each recognized independently, or at least having independent classification result on the output of the recognition system. When performing recognition results combination of such string objects additional problems arise -- such as the problems of ambiguous per-character segmentation, which leads to the combination of variable-length strings, etc.

To solve such problem a ROVER method (Recognizer Output Voting Error Reduction) can be used. This method was initially developed for improving speech recognition by combining multiple recognition results produced by different systems \cite{659110}. Its principal operation consists of two stages. On the first stage all recognition results are aligned in a single word transition network. In the case of text strings this could be done by inserting ``empty'' characters between some pairs of symbols in an optimal way. On the second stage a voting procedure is used to select the best recognition result for each component. The ROVER method can be generalized to the case where single component recognition result is not a class label, but rather a distribution of membership estimations \cite{vestnik_integration}. In such case to compute an optimal alignment a metric must be defined on the set of single component recognition results, including the ``empty'' character.

If a predictor could be constructed such that it would be possible to estimate validity of the recognition result, such predictor can be used for weighting the per-frame results in the combination. This could include zero-valued weights for rejecting some of the per-frame results which could spoil the overall integrated result, or to select and combine only the few ``best'' results. The question, however, arises -- which predictor to use to maximize the quality of the final result.

In this paper we propose two variants for weighting the per-frame recognition results before their combination. The weighting model is presented in section \ref{sec:model}, which unites the approaches of weighted combination and selection of best result, and the two weighting functions are introduced. Section \ref{sec:experiments} contains experimental evaluation of the proposed criteria.

\section{Weighting model and approaches} \label{sec:model}

\subsection{Constructing a weighting model}

If an individual frame was recognized with a bad precision (e.g. due to text field location or segmentation errors, low quality of an input image, etc.), and its recognition result is later used to produce a combined video sequence result, the final result may significantly degrade. If a method of prior estimation or individual result is accessible, these estimations could be used to introduce ordering of the frame results by its ``significance'', however another question arises -- which strategy is more appropriate, a weighted combination of all individual results, or a selection of a few best ones? In \cite{spie-lynchenko} this problem was considered in regards to a single character recognition, on two distinct data subsets: one containing localization and segmentation errors, and the other without such errors. The experiments showed that if localization and segmentation errors are present, the best result can be achieved using a combination of several most ``competent'' classifiers using a product rule or a voting procedure. With text field (or text string) recognition the problem is more complicated: there could be no single frame result on which all characters are recognized correctly or even visible, so one may expect that combination of all individual results might be more significant in the string recognition case.

Let us consider recognition of an object $x$ with a correct value $x^* \in \mathbb{X}$, where $\mathbb{X}$ is the set of all possible recognition results, in a sequence of video frames $I_1(x), \ldots, I_N(x)$, with $I_i(x) \in \mathbb{I}$, where $\mathbb{I}$ is the set of images. After processing each frame $I_i(x)$, a frame recognition result $x_i \in \mathbb{X}$ is obtained, and the quality of the result can be defined as a distance $\rho(x_i, x^*)$ according to some metric $\rho: \mathbb{X} \times \mathbb{X} \rightarrow \mathbb{R}^+_0$. An integration function $R^{(N)}:\mathbb{X}^N \times (\mathbb{R}^+_0)^N \rightarrow \mathbb{X}$ is defined, which takes a sequence of frame recognition results and their associated weights. A ``weight'' of the recognition result is a non-negative value which signifies the ``importance'' of the obtained result. With a fixed frame recognition module, and with a fixed integration function $R^{(N)}$, the task is to assign weights $w_1, w_2, \ldots, w_N$ to the frame recognition results $x_1, x_2, \ldots, x_N$ such that to minimize the expected quality of the integrated result $\rho(R^{(N)}(x_1, \ldots, x_N, w_1, \ldots, w_N), x^*)$. We assume here that if $w_i = 0$ then the frame result $x_i$ does not contribute to the integrated result at all.

Let us define a base weighting function $w:\mathbb{I} \times \mathbb{X} \rightarrow \mathbb{R}^+_0$ which takes as an input an image of an object and its recognition result, and produces a weight value, signifying the prior estimation of the ``quality'' of the image and/or its recognition result. Consider an ordering $\pi \in S_N$ of the frame recognition results in descending order of the corresponding values of a base weighting function: $\pi(i) < \pi(j) \Leftrightarrow w(I_i(x), x_i) \ge w(I_j(x), x_j)$, and a threshold parameter $t\in\{1, \ldots, N\}$. Using $\pi$ and $t$ we can construct a general weighting model as follows:
\begin{equation}
\label{eq:weighting_model}
w^{(t)}_i = \left\{\begin{aligned}
w(I_i(x), x_i), &\quad \text{if } \pi(i) \le t,\\
0, &\quad \text{if } \pi(i) > t
\end{aligned}\right..
\end{equation}

The weighting model \eqref{eq:weighting_model} can be used to generalized both the full weighted combination of all input samples (with $t = N$), the selection of the single best result (with $t = 1$ and with the $w$ determining the ``quality'' of the result) and intermediate systems with weighted combination of a few best frame results. Given such weighting model, the task is now to determine which base weighting function $w$ to use.

\subsection{Base weighting functions}

In the scope of this work, two criteria were evaluated as a base weighting function $w$. The first is a focus estimation $F(I(x))$ of an input image $I(x)$, calculated using an algorithm proposed in \cite{ecms-reducing-overconfidence}. This criterion was also used to control the input frame quality in video stream document recognition systems \cite{chernov2019application}. First, absolute values for vertical, horizontal, and diagonal image gradients are calculated:
\begin{equation}
\label{eq:gradients_calculation}
G^V_{i,j}(I)=| I_{i+1,j}-I_{i,j} |;~~
G^H_{i,j}(I)=| I_{i,j+1}-I_{i,j} |;~~
G^{D1}_{i,j}(I) = \cfrac{1}{\sqrt{2}}| I_{i+1,j+1}-I_{i,j} |;~~
G^{D2}_{i,j}(I) = \cfrac{1}{\sqrt{2}}|I_{i,j+1}-I_{i+1,j}|,
\end{equation}
where $I_{i,j}$ -- intensity value of the pixel with coordinates $(i, j)$ of an image $I$. Second, the focus estimation is calculated as a minimal $0.95$-quantile for obtained gradient values:
\begin{equation}
\label{eq:focus_estimation}
F(I) = \min\{q(G^V(I)), q(G^H(I)), q(G^{D1}(I)), q(G^{D2}(I))\},
\end{equation}
where $q(G)$ is a $0.95$-quantile of a gradient image $G$.  

The second weighting criterion used for classification quality estimation was an a-posteriori recognition confidence value $Q(x)$. The string object recognition result $x$ with length $M$ can be represented as an alternatives matrix:
\begin{equation}
\label{eq:alternatives_matrix}
x=\left(\begin{array}{ccc}
(q_{11}, c_{11})  & \cdots & (q_{M1}, c_{M1})\\
\vdots  & \ddots & \vdots\\
(q_{1K}, c_{1K})  & \cdots & (q_{MK}, c_{MK})
\end{array}\right),~~~c_{ij} \in C,~~~q_{ij} \in [0, 1],~~~\forall i\in\{1,\ldots,M\}: \sum\limits_{j=1}^K q_{ij} = 1,
\end{equation}
where $C$ is a set of all possible character alternatives, $card(C)=K$, and $q_{ij}$ is a membership estimation of the $i$-th element of the string object to a character class $c_{ij}\in C$.  The confidence value $Q(x)$ was calculated as a minimal value of the highest membership estimation across all string character classification results:
\begin{equation}
\label{eq:string_confidence_estimation}
Q(x)=\min\limits_{i=1}^M\left\{\max\limits_{j=1}^K q_{ij}\right\}.
\end{equation}

Note that the confidence value $Q(x)$ is calculated only using the independent per-frame recognition result and is based on individual character classification estimations. Further study of confidence-based weighting criteria may include more careful analysis of the changes in estimation distributions in a video stream \cite{janiszewski-modelling-the-flow}. 

\begin{figure}[ht!]
 \centering
 \includegraphics[width=0.8\textwidth]{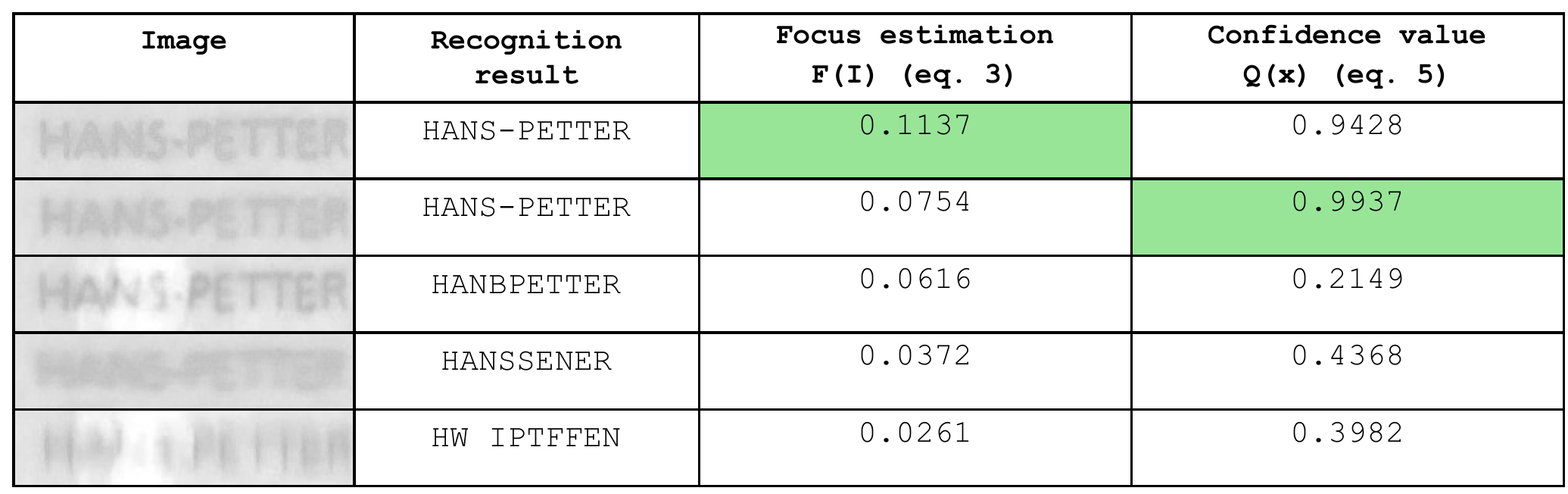}
 \caption{Illustration of base weighting functions. Images are taken from MIDV-500 dataset \cite{midv500-arxiv}}
 \label{fig:focus_illustration}
\end{figure}

Figure \ref{fig:focus_illustration} illustrates text field images, their recognition results, and corresponding weight values according to the two proposed criteria.

\subsection{Comparison methodology}

Per-frame recognition of text fields with subsequent combination of the obtained values in a single recognition result can be regarded as an anytime algorithm \cite{Zilberstein_1996} -- an iterative process which yields better results over time and can be stopped at any time with the best result available currently. A useful method to compare realizations of anytime algorithms is to use expected performance profiles -- graphical plots which show dependence of the expected result quality on the time required to obtain it. In the case of recognition results combination such performance profiles correspond to plots of mean distance between the obtained integrated results and ground truth against the length of the integrated clip prefix. In section \ref{sec:experiments} expected performance profiles will be presented for recognition results combination using different weighting methods.

\section{Experimental evaluation} \label{sec:experiments}

The constructed methods of weighted combination were tested on a MIDV-500 dataset \cite{midv500-arxiv}, which consists of 500 video sequences, each with 30 frames, of hand-held captured ID documents using mobile devices. Only the frames on which the document is fully visible were considered (if that lead to a reduction of the number of frames in a sequence, the remaining frames were repeated in a cycle until the length of each clip returned to 30 frames). From each image the text fields were cropped using a ground truth coordinates provided with the dataset. Four field groups were evaluated: numeric dates, document numbers, Latin name components, and MRZ lines. Each text field image was recognized using a text line recognition subsystem of Smart IDReader document recognition system \cite{SMART_IDREADER_ICDAR}. The comparison with correct text field values was case-insensitive, and the letter ``\verb|O|'' was treated as identical to the digit ``\verb|0|''. Normalized Generalized Levenshtein Distance \cite{ngld_yujian} was used as a metric function $\rho(x_i, x^*)$, as in work \cite{vestnik_integration}.

In terms of the weighting model \eqref{eq:weighting_model} five approaches to weighted combination were considered given a base weighting functions: combination without weighting (i.e. using a constant value as a base weighting function and a threshold parameter $t=N$), choosing the best (threshold parameter $t=1$) and weighted combinations of the 3 best (threshold parameter $t=3$), of the best 50\% (threshold parameter $t=N/2$), and of all frames (threshold parameter $t=N$).

\begin{figure}[ht!]
 \centering
 \includegraphics[width=0.95\textwidth]{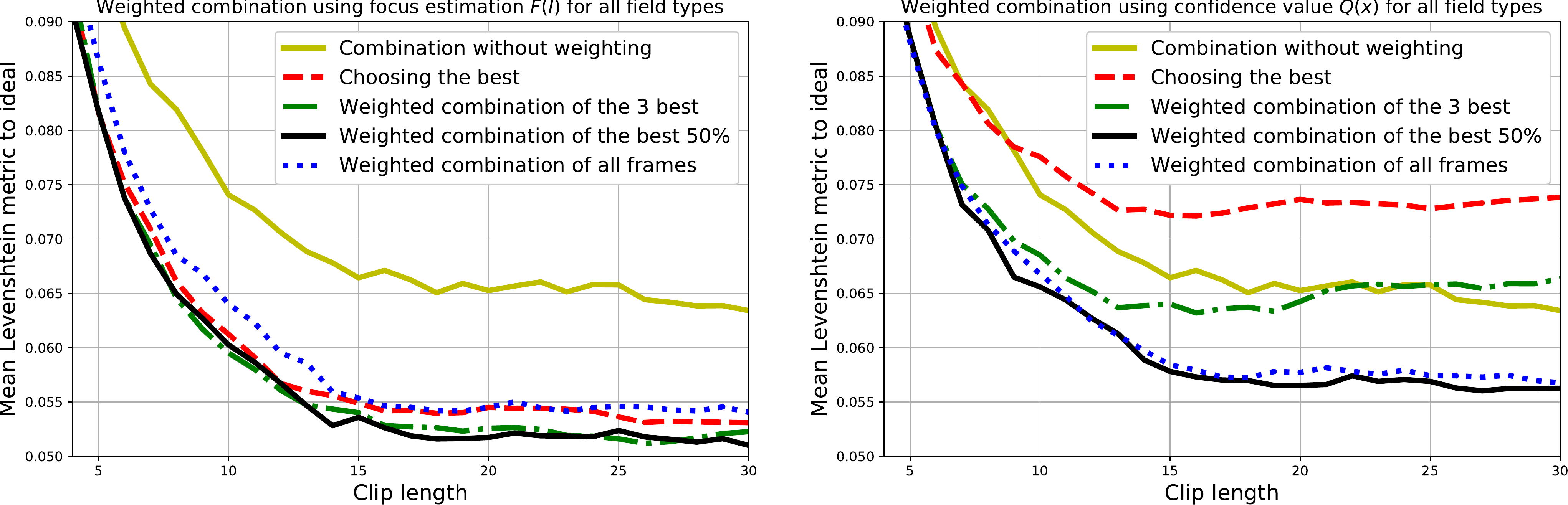}
 \caption{Performance profiles for weighted combination for all field types based on focus estimation (left) and confidence value (right)}
 \label{fig:graphics_all_types}
\end{figure}

Figure \ref{fig:graphics_all_types} shows the performance profiles for various approaches to weighted combination using a focus estimation \eqref{eq:focus_estimation} and a recognition result confidence value \eqref{eq:string_confidence_estimation}, as measure on all analyzed field groups of the MIDV-500 dataset. 

It can be seen that on the analyzed dataset and based on confidence value weighting function the selection of a single best result yields the worst result. Using both weighting functions the weighted combination of best 50\% frames yields the best result, and the result is notably better when compared with unweighted combination.

\begin{figure}[ht!]
 \centering
 \includegraphics[width=0.45\textwidth]{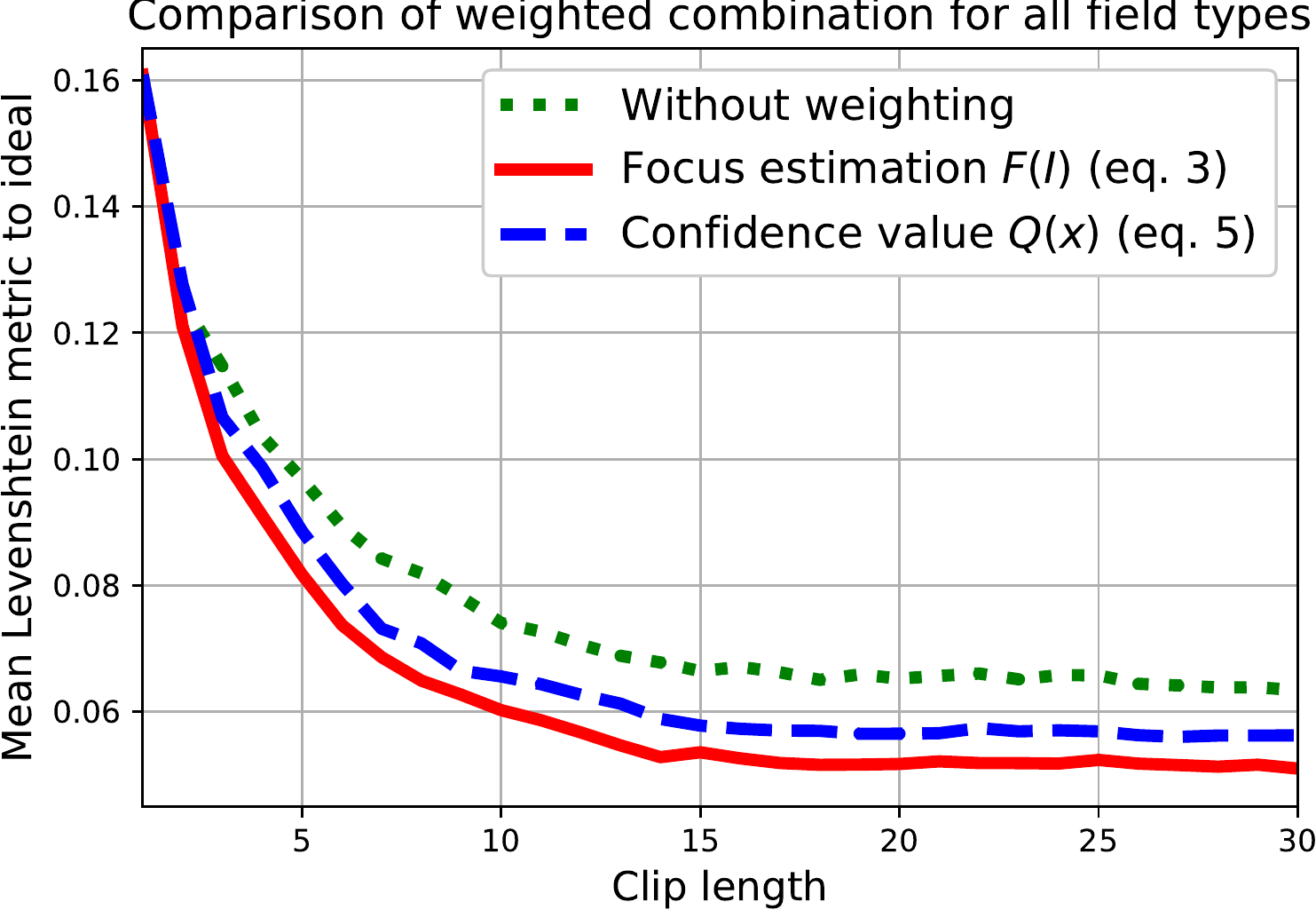}
 \caption{Comparison of weighted combination of all frames using focus estimation and confidence value for all field types}
 \label{fig:comparative_graphics_all_fields}
\end{figure}

Figure \ref{fig:comparative_graphics_all_fields} shows the performance profiles for weighted combination of the best 50\% frames (determined as the best strategy from the considered five alternatives) for focus estimation and confidence value, in order to compare the criteria. It can be observed that with any length of the integrated clip prefix the focus estimation $F(I)$~\eqref{eq:focus_estimation} performs better than the confidence value.

Table \ref{tbl:results_for_diff_criteria} shows mean Levenshtein metric to ideal for different weighting criteria, separately for the four analyzed field groups and the total result for all analyzed fields. Thus the best weighted combination strategy for all evaluated weighting functions was to use the best 50\% input frames. The weighted combination outperforms the one without assigning per-frame sample weights, and the weighting based on focus estimation of field image \eqref{eq:focus_estimation} turned out to be the most effective for generic field groups.

\begin{table}
    \centering
	\caption{Mean Levenshtein metric to ideal for different weighting criteria}
	\label{tbl:results_for_diff_criteria}       
	\resizebox{0.9\textwidth}{!}{
		\begin{tabular}{|l|l|l|l|l|l|}
			\hline
			\parbox{0.25\textwidth}{Weighting criterion} &
			\parbox{0.11\columnwidth}{Doc. number} & 
			\parbox{0.11\columnwidth}{Date} & 
			\parbox{0.11\columnwidth}{Latin} & 
			\parbox{0.11\columnwidth}{MRZ}  &
			\parbox{0.11\columnwidth}{All fields}

			\\
			\hline
			\hline
			
			\parbox{0.25\columnwidth}{No weighting} & $0.0605$
			& $0.0657$ & $0.0735$ & $\mathbf{0.0333}$ & $0.0634$

			\\
			\hline
			
			\parbox{0.25\columnwidth}{Focus estimation $F(I)$ \eqref{eq:focus_estimation}} & $\mathbf{0.0431}$ & $\mathbf{0.0525}$ & $\mathbf{0.0599}$ & $0.0351$ & $\mathbf{0.0510}$

			\\
			\hline
			
			\parbox{0.25\columnwidth}{Confidence value $Q(x)$ \eqref{eq:string_confidence_estimation}} & $0.0492$ & $0.0581$ & $0.0682$ & $0.0336$ & $0.0568$ 
			\\
			
			\hline
		\end{tabular}
	}
\end{table}

\section{Conclusion}

In this paper we considered the problem of combining multiple recognition results in relation to the task of improving the quality of text fields recognition in a video stream. Evaluations conducted on a MIDV-500 video data set shows, that the use of weighted combination allows improve text recognition precision in a video stream.

Proposed model generalizes the combination and selection of the best result by a pre-defined criterion. Two different weighting criteria -- the string recognition result confidence score and the input image focus score -- were investigated for usage with the weighting model. The input image focus score weighting criteria yield consistently higher total precision in the performed experiments.

In view of this result the weighted combination should be researched further, and other weighting criteria might be considered, such as ones based on a field language model, which does not depend on the particular recognition method.

\acknowledgements

This work is partially financially supported by Russian Foundation for Basic Research (projects 17-29-03236, 18-07-01387). 

\bibliographystyle{spiebib}
\bibliography{bibliography}

\begin{thebibliography}{10}

\bibitem{a_survey_of_ocr_applications}
A. Singh, K. Bacchuwar, and A. Bhasin, ``A survey of {OCR} applications,''
  {International Journal of Machine Learning and Computing ({IJMLC})}~{\bf
  2}(3),  314--318 (2012).
\newblock \doi{10.7763/IJMLC.2012.V2.137}.

\bibitem{a_survey_on_ocr_arxiv}
N. Islam, Z. Islam, and N. Noor, ``A survey on optical character recognition
  system,'' {arXiv.1710.05703}  (2017).

\bibitem{8342338}
M.~R. {Soheili}, M.~R. {Yousefi}, E. {Kabir}, and D. {Stricker}, ``Merging
  clustering and classification results for whole book recognition,'' in 2017
  10th Iranian Conference on Machine Vision and Image Processing
  (MVIP){\nolinebreak\hspace{0.1em}},   134--138 (2017).

\bibitem{icmv-jabnoun}
H. Jabnoun, F. Benzarti, and H. Amiri, ``A new method for text detection and
  recognition in indoor scene for assisting blind people,'' in Proc. SPIE (ICMV
  2016){\nolinebreak\hspace{0.1em}},   {\bf 10341} (2017).
\newblock \doi{10.1117/12.2268399}.

\bibitem{arl-small-scale-cameras}
V.~V. Arlazarov, A. Zhukovsky, V. Krivtsov, D. Nikolaev, and D. Polevoy,
  ``Analysis of using stationary and mobile small-scale digital cameras for
  documents recognition,'' {Information Technologies and Computing Systems}
  (3),  71--81 (2014).
\newblock (in Russian).

\bibitem{7881422}
O. {Alaql}, K. {Ghazinour}, and C.~C. {Lu}, ``Classification of image
  distortions for image quality assessment,'' in 2016 International Conference
  on Computational Science and Computational Intelligence
  (CSCI){\nolinebreak\hspace{0.1em}},   653--658 (2016).
\newblock \doi{10.1109/CSCI.2016.0129}.

\bibitem{6507950}
K. {Anantharajah}, S. {Denman}, S. {Sridharan}, C. {Fookes}, and D.
  {Tjondronegoro}, ``Quality based frame selection for video face
  recognition,'' in 2012 6th International Conference on Signal Processing and
  Communication Systems{\nolinebreak\hspace{0.1em}},   1--5 (2012).
\newblock \doi{10.1109/ICSPCS.2012.6507950}.

\bibitem{6969211}
C. {Merino-Gracia} and M. {Mirmehdi}, ``Real-time text tracking in natural
  scenes,'' {IET Computer Vision}~{\bf 8}(6),  670--681 (2014).
\newblock \doi{10.1049/iet-cvi.2013.0217}.

\bibitem{efficient_video_scene_text_spotting_arxiv}
Z. Cheng, J. Lu, J. Xie, Y. Niu, S. Pu, and F. Wu, ``Efficient video scene text
  spotting: Unifying detection, tracking, and recognition,'' {arXiv.1903.03299}
   (2019).

\bibitem{YUE2016389}
L. Yue, H. Shen, J. Li, Q. Yuan, H. Zhang, and L. Zhang, ``Image
  super-resolution: The techniques, applications, and future,'' {Signal
  Processing}~{\bf 128},  389--408 (2016).
\newblock \doi{10.1016/j.sigpro.2016.05.002}.

\bibitem{5995614}
C. {Liu} and D. {Sun}, ``A {B}ayesian approach to adaptive video super
  resolution,'' in CVPR 2011{\nolinebreak\hspace{0.1em}},   209--216 (2011).
\newblock \doi{10.1109/CVPR.2011.5995614}.

\bibitem{vestnik_integration}
K. Bulatov, ``A method to reduce errors of string recognition based on
  combination of several recognition results with per-character alternatives,''
  {Bulletin of the South Ural State University. Ser. Mathematical Modelling,
  Programming~\&~Computer Software}~{\bf 12}(3),  74--88 (2019).
\newblock \doi{10.14529/mmp190307}.

\bibitem{chernov2019application}
T.~S. Chernov, S.~A. Ilyuhin, and V.~V. Arlazarov, ``Application of dynamic
  saliency maps to video stream recognition systems with image quality
  assessment,'' in Proc. SPIE (ICMV 2018){\nolinebreak\hspace{0.1em}},   {\bf
  11041}(110410T) (2019).
\newblock \doi{10.1117/12.2522768}.

\bibitem{ensemble_methods}
Z.-H. Zhou,  {Ensemble Methods: Foundations and
  Algorithms}{\nolinebreak\hspace{0.1em}}, Chapman and Hall/CRC, New York,
  1~ed. (2012).
\newblock \doi{10.1201/b12207}.

\bibitem{1688199}
R. {Polikar}, ``Ensemble based systems in decision making,'' {IEEE Circuits and
  Systems Magazine}~{\bf 6}(3),  21--45 (2006).
\newblock \doi{10.1109/MCAS.2006.1688199}.

\bibitem{659110}
J.~G. {Fiscus}, ``A post-processing system to yield reduced word error rates:
  {R}ecognizer {O}utput {V}oting {E}rror {R}eduction ({ROVER}),'' in 1997 IEEE
  Workshop on Automatic Speech Recognition and Understanding
  Proceedings{\nolinebreak\hspace{0.1em}},   347--354 (1997).
\newblock \doi{10.1109/ASRU.1997.659110}.

\bibitem{spie-lynchenko}
K. Bulatov, A. Lynchenko, and V. Krivtsov, ``Optimal frame-by-frame result
  combination strategy for {OCR} in video stream,'' in Proc. SPIE (ICMV
  2017){\nolinebreak\hspace{0.1em}},   {\bf 10696}(106961Z) (2018).
\newblock \doi{10.1117/12.2310139}.

\bibitem{ecms-reducing-overconfidence}
K. B.~Bulatov and D. V.~Polevoy, ``Reducing overconfidence in neural networks
  by dynamic variation of recognizer relevance,'' in Proceedings - 29th
  European Conference on Modelling and Simulation, ECMS
  2015{\nolinebreak\hspace{0.1em}},   488--491 (2015).
\newblock \doi{10.7148/2015-0488}.

\bibitem{janiszewski-modelling-the-flow}
V.~V. Arlazarov, O. Slavin, A. Uskov, and I. Janiszewski, ``Modelling the flow
  of character recognition results in video stream,'' {Bulletin of the South
  Ural State University. Ser. Mathematical Modelling, Programming~\&~Computer
  Software}~{\bf 11},  14--28 (2018).
\newblock \doi{10.14529/mmp180202}.

\bibitem{midv500-arxiv}
V.~V. {Arlazarov}, K. {Bulatov}, T. {Chernov}, and V.~L. {Arlazarov}, ``{A
  Dataset for Identity Documents Analysis and Recognition on Mobile Devices in
  Video Stream},'' {arXiv.1807.05786}  (2018).

\bibitem{Zilberstein_1996}
S. Zilberstein, ``Using anytime algorithms in intelligent systems,'' {AI
  Magazine}~{\bf 17}(3),  73--83 (1996).
\newblock \doi{10.1609/aimag.v17i3.1232}.

\bibitem{SMART_IDREADER_ICDAR}
K. {Bulatov}, V.~V. {Arlazarov}, T. {Chernov}, O. {Slavin}, and D. {Nikolaev},
  ``Smart {IDR}eader: Document recognition in video stream,'' in 14th
  International Conference on Document Analysis and Recognition
  ({ICDAR}){\nolinebreak\hspace{0.1em}},   {\bf 6},  39--44, IEEE (2017).
\newblock \doi{10.1109/ICDAR.2017.347}.

\bibitem{ngld_yujian}
L. Yujian and L. Bo, ``A normalized levenshtein distance metric,'' {IEEE
  Transactions on Pattern Analysis and Machine Intelligence}~{\bf 29}(6),
  1091--1095 (2007).
\newblock \doi{10.1109/TPAMI.2007.1078}.

\end{thebibliography}

\clearpage

\end{document}